\documentclass[10pt,twocolumn,letterpaper]{article}

\usepackage[pagenumbers]{cvpr}

\usepackage{graphicx}
\usepackage{amsmath}
\usepackage{amssymb}
\usepackage{booktabs}
\usepackage{pifont}
\usepackage{multirow}
\usepackage{soul}
\usepackage{array}
\usepackage{rotating}
\usepackage[para]{footmisc}
\usepackage[accsupp]{axessibility}  
\newcolumntype{P}[1]{>{\centering\arraybackslash}p{#1}}

\usepackage[pagebackref,breaklinks,colorlinks]{hyperref}

\newcommand*\samethanks[1][\value{footnote}]{\footnotemark[#1]}

\usepackage[capitalize]{cleveref}
\crefname{section}{Sec.}{Secs.}
\Crefname{section}{Section}{Sections}
\Crefname{table}{Table}{Tables}
\crefname{table}{Tab.}{Tabs.}

\def\cvprPaperID{3051} 
\def\confName{CVPR}
\def\confYear{2022}

\newtoggle{arxiv}
\newtoggle{selfappendix}

\togglefalse{arxiv}
\togglefalse{selfappendix}

\toggletrue{arxiv}

\begin{document}

\title{AdaInt: Learning Adaptive Intervals for 3D Lookup Tables \\on Real-time Image Enhancement}

\author{Canqian Yang$^{1,4}$\thanks{Equal Contribution\hspace*{-1.8em}},\quad
Meiguang Jin$^2$\samethanks,\quad
Xu Jia$^3$,\quad
Yi Xu$^{1,4}$\thanks{Corresponding Author\newline\hspace*{1.8em}Work partially done during an internship of C. Yang at Alibaba Group.},\quad
Ying Chen$^2$\\
$^1$MoE Key Lab of Artificial Intelligence, AI Institute, Shanghai Jiao Tong University\quad $^2$Alibaba Group\\
$^3$Dalian University of Technology\quad $^4$Chongqing Research Institute, Shanghai Jiao Tong University\\
{\tt\small \{charles.young, xuyi\}@sjtu.edu.cn}\\
{\tt\small xjia@dlut.edu.cn\quad \{meiguang.jmg, chenying.ailab\}@alibaba-inc.com}
}

\maketitle

\begin{abstract}

The 3D Lookup Table (3D LUT) is a highly-efficient tool for real-time image enhancement tasks, which models a non-linear 3D color transform by sparsely sampling it into a discretized 3D lattice. Previous works have made efforts to learn image-adaptive output color values of LUTs for flexible enhancement but neglect the importance of sampling strategy. They adopt a sub-optimal uniform sampling point allocation, limiting the expressiveness of the learned LUTs since the (tri-)linear interpolation between uniform sampling points in the LUT transform might fail to model local non-linearities of the color transform. Focusing on this problem, we present \textbf{AdaInt} (Adaptive Intervals Learning), a novel mechanism to achieve a more flexible sampling point allocation by adaptively learning the non-uniform sampling intervals in the 3D color space. In this way, a 3D LUT can increase its capability by conducting dense sampling in color ranges requiring highly non-linear transforms and sparse sampling for near-linear transforms. The proposed AdaInt could be implemented as a compact and efficient plug-and-play module for a 3D LUT-based method. To enable the end-to-end learning of AdaInt, we design a novel differentiable operator called \textbf{AiLUT-Transform} (Adaptive Interval LUT Transform) to locate input colors in the non-uniform 3D LUT and provide gradients to the sampling intervals. Experiments demonstrate that methods equipped with AdaInt can achieve state-of-the-art performance on two public benchmark datasets with a negligible overhead increase. Our source code is available at \url{https://github.com/ImCharlesY/AdaInt}.

\vspace{-0.4cm}

\end{abstract}

\section{Introduction}
\label{sec:intro}

\begin{figure}[ht]
    \centering
    \includegraphics[width=\linewidth]{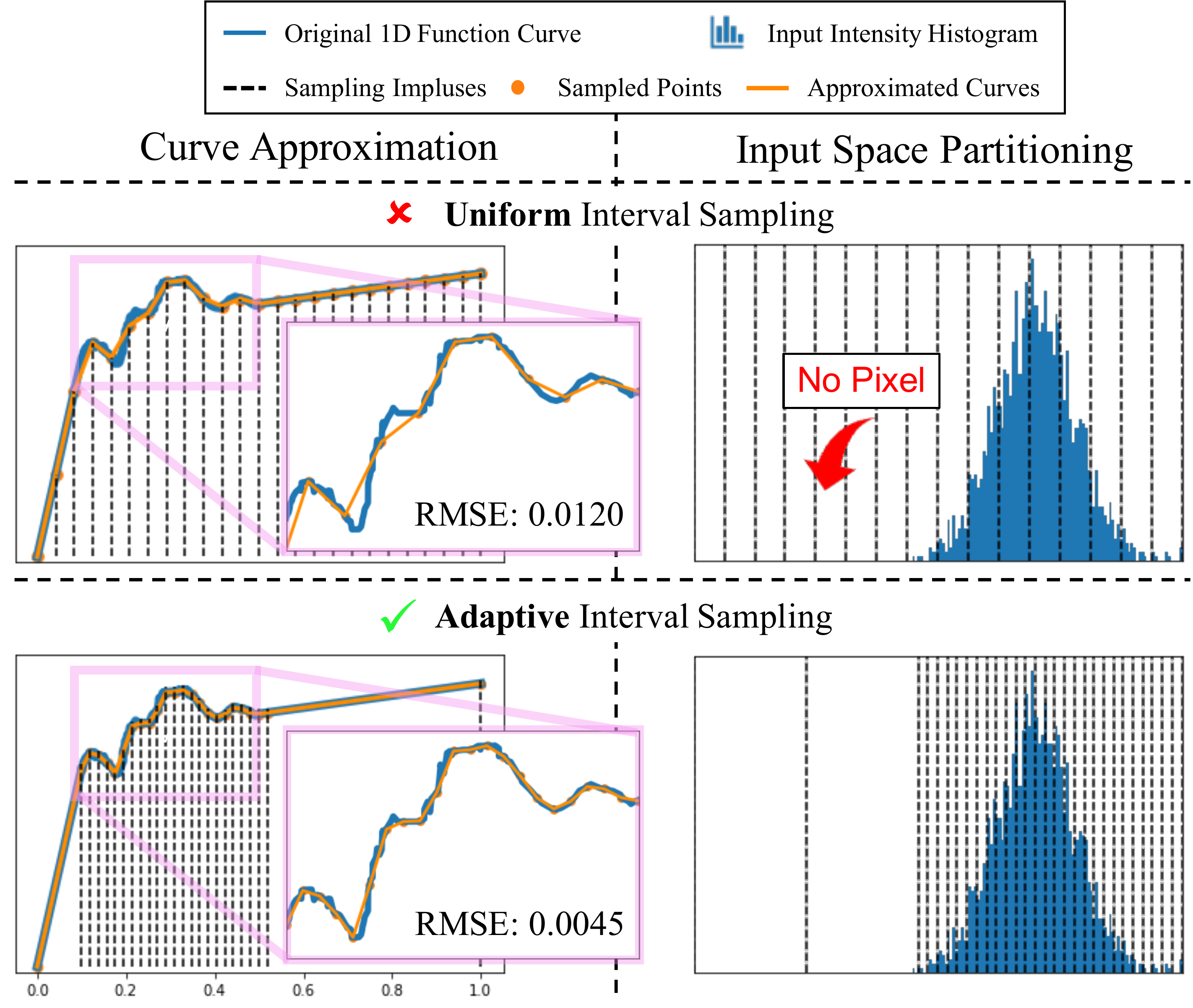}
    \caption{Comparison between uniform and non-uniform sampling on curve approximation and space partitioning. The illustration is given in 1D but can be easily extended to 3D.}
    \label{fig:motivation}
    \vspace{-0.6cm}
\end{figure}

Recent advances in machine learning techniques remarkably boosted the performance of automatic photo enhancement methods~\cite{TOG16Auto, CGF12content, ECCV12context, CVPR18DPE, ACMMM18AD, CVPR18Dark}, aiming to replace a sequence of meticulously-designed operations~\cite{GP98Exposure1,ECCV12Exposure2,TCE97contrast,TIP18learning,ACMSIG08display,TPAMI12new} in the camera imaging pipeline~\cite{ECCV16software} for enhanced visual quality. However, these methods suffer from heavy computational burdens due to complicated optimization processes~\cite{CVPR08bayesian, ICCVW17new, CVPR16weighted, CGF12content, ECCV12context} or neural architecture designs~\cite{CVPR18DPE, ACMMM18AD, BMVC18Lowlight, CVPR18Dark}. In fact, most of the commonly-used enhancement operations are pixel-independent, as revisited in~\cite{ECCV20CSRNet}. Their total effect is approximately equivalent to a 3D color transform function ($\mathbb{R}^3\rightarrow\mathbb{R}^3$) that maps an input color point to another one in or across the color spaces.
One can adopt a multi-layer perceptron (MLP) to design such a transform \cite{ECCV20CSRNet} but requires a cascade of several linear and nonlinear sub-operations to increase the model capability.
To overcome the computational complexity of a series of sub-operations in the transform, the 3D lookup table (LUT) is a promising data structure to conduct efficient mapping by sparsely sampling a range of input values and storing the corresponding output values in a 3D lattice. The non-linearities in the transform are typically approximated by a set of (tri-)linear interpolation functions distributed in the lattice cells. Since a LUT can transform images using only memory access and interpolation operations, it shows an advantage of high efficiency and practicality.

Previous works~\cite{TPAMI203DLUT,ICCV21SALUT} have made efforts to learn an image-adaptive LUT, mimicking the underlying optimal color transform with adaption to extensively varied image content. These methods embody the image-adaptiveness of the 3D LUTs only in the output color values, which are automatically learned by neural networks~\cite{TPAMI203DLUT, ICCV21SALUT}. However, they conduct sampling with equal intervals, not considering the adaption of sampling point density to image contents. It results in a sub-optimal sampling point allocation, limiting the expressiveness of the LUTs to model local non-linearities. Specifically, input pixels with similar color values but requiring highly non-linear contrast stretching (\textit{e.g.}, enhancement on low-light texture regions) are possibly compressed into the same lattice cell, which ends up producing linear-scaling results. The reasons lie in limited sampling points and (tri-)linear interpolation in the LUT transform. As depicted in the left part of \Cref{fig:motivation}, for example, a uniform spacing undersamples a color range where the transform exhibits high curvature, resulting in distortion of the non-linearities in the transform. Ideally, increasing the number of sampling points might mitigate the issues but will significantly increase the overhead of the 3D LUT. Besides, it would also aggravate the oversampling in color space where few pixels fall into, causing waste in the LUT capacity, as shown in the right part of \Cref{fig:motivation}.

\begin{figure*}[t]
    \centering
    \includegraphics[width=\linewidth]{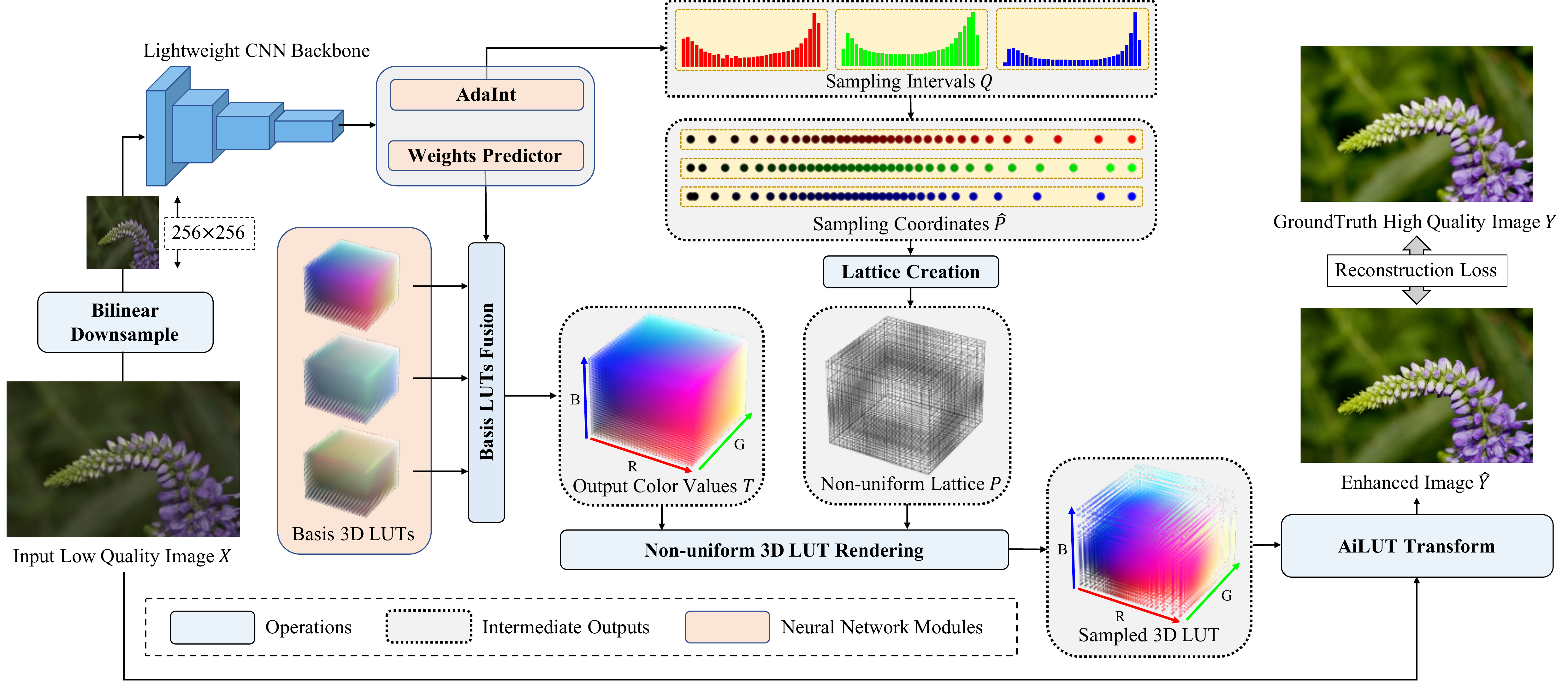}
    \caption{Framework of the proposed method. Our method employs a CNN model on the down-sampled version of the input image to simultaneously predict two fundamental components of an image-adaptive 3D LUT – the sampling coordinates and output values. These two components construct a dedicated 3D LUT in a non-uniform layout via the lattice creation and rendering processes. The input image of the original resolution can be afterward transformed by the predicted 3D LUT efficiently via a designed novel AiLUT-Transform. The overall framework can be trained under the supervision of the groundtruth images in an end-to-end manner. Best viewed in color.}
    \label{fig:framework}
    \vspace{-0.4cm}
\end{figure*}

To achieve a better tradeoff between effectiveness and efficiency when given limited sampling points, we develop a novel deep learning-based approach to adjust the layout of the 3D lattice by dynamically learning the non-uniform sampling intervals. This idea is encapsulated into a compact network module called \textbf{AdaInt}, that can adaptively allocate more sampling points to color ranges requiring highly non-linear transforms and reduce redundant sampling point quota for strongly linear ranges. As illustrated in \Cref{fig:framework}, with the incorporation of AdaInt, a lightweight convolutional neural network (CNN) takes a down-sampled image as input to simultaneously predict two components of a dedicated 3D LUT – the non-uniform sampling coordinates and the corresponding output color values. These two components are combined to compose an image-adaptive 3D LUT that transforms the original image via a novel differentiable operator called AiLUT-Transform, which can provide gradients to AdaInt for end-to-end learning. This operator is essential for locating input colors in a non-uniform 3D lattice by introducing a low-complexity binary search into the lookup procedure of a LUT transform. Therefore, our method could be a plug-and-play module for 3D LUTs and still presents high efficiency.

The main contributions of this paper are three-fold:
(1) We view the learning of 3D LUTs from the viewpoint of sampling and point out the importance of the sampling strategy for modeling color transforms with higher non-linear capability.
(2) We present a novel AdaInt module and the corresponding AiLUT-Transform operator to enable the adaptive learning of a 3D LUT with a non-uniform layout.
(3) We demonstrate the effectiveness and efficiency of our method on two large-scale publicly available datasets.

\section{Related Works}
\label{sec:related}

\subsection{Photo Enhancement Methods}

Recent advances in learning-based image enhancement methods can be roughly divided into two categories. The first paradigm~\cite{CVPR18DPE,CVPR20DeepLPF,ACMMM18AD, ACMMM19Lowlight,BMVC18Lowlight,CVPR18Dark} directly learns the dense end-to-end mapping via fully convolutional networks~\cite{FCN}. While this line of works can achieve promising results, they usually suffer from heavy computational and memory burdens, limiting their practicalities.
The second paradigm simultaneously leverages the strong fitting abilities of deep learning and the high efficiency of traditional physical models. This line of studies commonly transfers the heavy CNN dense inference into light physical model parameter prediction. The physical models are then used to enhance original images efficiently. The frequently used physical models include affine color transforms~\cite{TOG17HDRNet,CVPR19UPE,WACV20supervised,ACCV20color}, mapping curves (which can be viewed as 1D LUTs)~\cite{ICPR21curl,ICCV21Star,ICCV21RCT,ECCV20Global,Arxiv20flexible,CVPR20zero,CVPR18DisRec}, multi-layer perceptrons (MLPs)~\cite{ECCV20CSRNet} and 3D color LUTs~\cite{TPAMI203DLUT,ICCV21SALUT}. Among them, 3D LUT is the most promising one due to its faster speed than MLPs, along with the stronger capability than affine transforms and mapping curves. The works most related to ours are~\cite{TPAMI203DLUT,ICCV21SALUT}, which also learn image-adaptive 3D LUTs for enhancing images in real-time. However, they learn 3D LUTs in a uniform layout without considering the image-adaptiveness of the sampling strategy, which restricts their ability to model non-linear color transform.

\subsection{Non-uniform Sampling}

Non-uniform sampling strategies have been extensively investigated in 3D shape recognition such as meshes~\cite{ICCV19mesh}, point clouds~\cite{CVPR17pointnet}, and implicit function fields~\cite{CVPR19occupancy} due to their higher efficiency and expressiveness compared to regular grids. For 2D image analysis, while the dominant paradigm is computation on regular 2D grids, recent works have made attempts to the non-uniform sampling of the input images~\cite{ICCV19efficient}, output images~\cite{CVPR20PointRend}, feature maps~\cite{ICCV17DCN}, and convolution filters~\cite{ICLR19DKN}. These works showed that an adaptive sampling strategy enables a high-quality representation using fewer sampling points.
Non-uniform layouts have also emerged in traditional LUT implementation~\cite{ECCV12Nonuniform,TIP97Seq,CIC08SSD}. However, these works focus on another task, \textit{lattice regression}~\cite{NIPS09Lattice}, aiming to fit a \textit{known} color transform into a \textit{static} 3D LUT and \textit{repeat} the transform during inference. The non-uniform layouts are introduced as an alternative way to reduce estimation errors. However, these methods are not flexible and intelligent as the estimated LUT is fixed and cannot adapt to new samples. Instead, our work learns the non-uniform 3D LUTs based on the content of every single image for more intelligent enhancement.

\label{sec:related-works}

\section{Method}
\label{sec:method}

\subsection{Preliminary: 3D Lookup Tables}
\label{sec:method-3dlut}

In this paper, we view a 3D LUT as a discrete sampling of a complete 3D color transform function. The sampled results are stored in a 3D lattice of output color values $T=\{(T_{r,(i,j,k)}, T_{g,(i,j,k)}, T_{b,(i,j,k)} )\}_{i,j,k\in\mathbb{I}_0^{N_s-1}}$ that can be queried by sets of input color coordinates $P=\{(P_{r,(i,j,k)}, P_{g,(i,j,k)}, P_{b,(i,j,k)}\}_{i,j,k\in\mathbb{I}_0^{N_s-1}}$, where $N_s$ is the number of sampling coordinates along each of three dimensions and $\mathbb{I}_0^{N_s-1}$ denotes the set of $\{0, 1, \dots, N_s-1\}$. Such a lattice defines a total of $N_s^3$ sampling points on the complete 3D color transform function. Once a 3D lattice is sampled, an input pixel looks up its nearest sampling points according to its color and computes its transformed output via interpolation (typically trilinear interpolation).

Due to the high efficiency and stability of 3D LUTs, previous methods~\cite{TPAMI203DLUT, ICCV21SALUT} have tried creating automatic image enhancement tools by learning image-adaptive 3D LUTs. They predict image-adaptive output values $T\in[0,1]^{3\times N_s\times N_s\times N_s}$ by learning several basis 3D LUTs and fusing them using image-dependent weights. These weights are predicted by a CNN model from the down-sampled input image, which significantly saves the computational cost (see the left part of \Cref{fig:framework}). However, these methods \textit{uniformly} discretize the 3D color space, not considering the image-adaptiveness of sampling coordinates $P\in[0,1]^{3\times N_s\times N_s\times N_s}$, making them suffer from sub-optimal sampling point allocation and limited LUT capability.

In this paper, we address the above issues by simultaneously learning the sampling coordinates and the corresponding output color values in an image-adaptive fashion. \Cref{fig:framework} shows an overview of the proposed framework. We directly follow the practice in~\cite{TPAMI203DLUT} to predict a set of candidate output color values $T$ due to its proven effectiveness. Suppose that $N_s$ sampling coordinates along each dimension and an input image $X\in[0,1]^{3\times H\times W}$ are given. The output color values of a LUT can be formulated as
\begin{equation}
    \label{eq:outval}
    T=h(f(X)),
\end{equation}
where $f$ is a function mapping an input image into a compact vector representation $E\in\mathbb{R}^F$. The function $h$ takes $E$ as input and predicts all output color values in $T$. Note that we encapsulate the idea of learning $M$ image-independent basis 3D LUTs and $M$ image-adaptive weights~\cite{TPAMI203DLUT} into a cascade of two mappings, denoted as $h: \mathbb{R}^F \xrightarrow[]{h_0} \mathbb{R}^M \xrightarrow[]{h_1}[0,1]^{3\times N_s\times N_s\times N_s}$, with the insight of using rank factorization to save parameters. The basis 3D LUTs are encoded as the parameters of $h_1$. Please refer to \iftoggle{arxiv}{\Cref{sec:suppl-arch}}{the supplementary materials} for more details. In the following section, we focus more on the learning of the sampling coordinates $P$. 

\subsection{Adaptive Intervals Learning (AdaInt)}
\label{sec:method-adaint}

Predicting the sampling color coordinates is equivalent to learning the placement of the sampling points in the 3D color space. Although the totally free sampling points placement provides high flexibility, it complicates the lookup procedure and increases the overhead significantly. To this end, we present a simple yet effective way to achieve the so-called \textit{constrained sampling point placement}. First, we assume that the three lattice dimensions are independent of each other during the lookup procedure. In this way, we can predict the sampling coordinates along each lattice dimension separably. Second, we reparameterize the sampling coordinates by the intervals between each adjacent pair of them. Therefore, by converting the learning goal from sampling coordinates to sampling intervals, we propose a novel image-adaptive constrained sampling point placement method, termed \textit{AdaInt}, which we illustrate in the following four steps.

\paragraph{Unnormalized Intervals Prediction}
Our method first predicts different sets of $N_s-1$ unnormalized sampling intervals for three lattice dimensions, thus producing a total of $3\times(N_s-1)$ values of intervals:
\begin{equation}
    \label{eq:adaint}
    \hat{Q}\in\mathbb{R}^{3\times(N_s-1)} = g(f(X)).
\end{equation}
In this work, we share the mapping $f$ between sampling points prediction and output values prediction. $g$ denotes a mapping of $\mathbb{R}^F \rightarrow \mathbb{R}^{3\times(N_s-1)}$. Please refer to \Cref{sec:exp-impl} for more implementation details.

\paragraph{Intervals Normalization}
Since the input and output spaces are normalized, the intervals for a given dimension should also spread out in the range of $[0, 1]$. In this work, we choose the \textit{softmax} function to get the normalized intervals $Q\in[0,1]^{3\times(N_s-1)} = \text{softmax}(\hat{Q}, \text{axis}=1)$ for convenience. The term "$\text{axis}=1$" indicates the normalization is performed on each of three color dimensions separably.

\paragraph{Intervals to Coordinates Conversion}
The sampling coordinates $\hat{P}\in[0,1]^{3\times N_s}$ are obtained by applying cumulative summation to $Q$ and prepending an origin to each lattice dimension, which can be formulated as:
$\hat{P} = [0_3^T; \text{cumsum}(Q, \text{axis}=1)]$, where $0_3$ is a 3-dimension zero vector, and the $[\cdot\ ;\ \cdot]$ symbol denotes the concatenation operation. The above operations guarantee the \textit{bounded} ($0 \leq \hat{P}_{c,i} \leq 1$, for $\forall c=r,g,b$ and $\forall i\in\mathbb{I}_0^{N_s-1}$) and the \textit{monotone increasing} properties ($\hat{P}_{c,i} \leq \hat{P}_{c,j}$, for $\forall c=r,g,b$, and $\forall i,j\in\mathbb{I}_0^{N_s-1}, i\leq j$) of the predicted sampling coordinates along each dimension, which significantly simplifies the lookup procedure to be presented in \Cref{sec:method-ailut-transform}.

\paragraph{Non-uniform 3D Lattice Construction}
The above $\hat{P}$ matrix indeed provides three $N_s$-dimension coordinate vectors for each lattice dimension, respectively. We can derive the 3D coordinates $P\in[0,1]^{3\times N_s\times N_s\times N_s}$ of the $N_s^3$ sampling points by calculating the n-ary Cartesian product ($\otimes$) over these $3$ coordinate vectors, \textit{i.e.},
$P = \hat{P}_r \otimes \hat{P}_g \otimes \hat{P}_b = \{(\hat{P}_{r,i}, \hat{P}_{g,j}, \hat{P}_{b,k})|i,j,k\in\mathbb{I}_0^{N_s-1}\}$.
These coordinates determine the vertex locations of a non-uniform 3D lattice. The final 3D LUT is easily constructed by assigning each output color value in $T$ to the corresponding vertex defined in $P$. Such a procedure can be vividly analogized to a rendering process, as illustrated in \Cref{fig:framework}.

\begin{figure}[t]
    \centering
    \includegraphics[width=\linewidth]{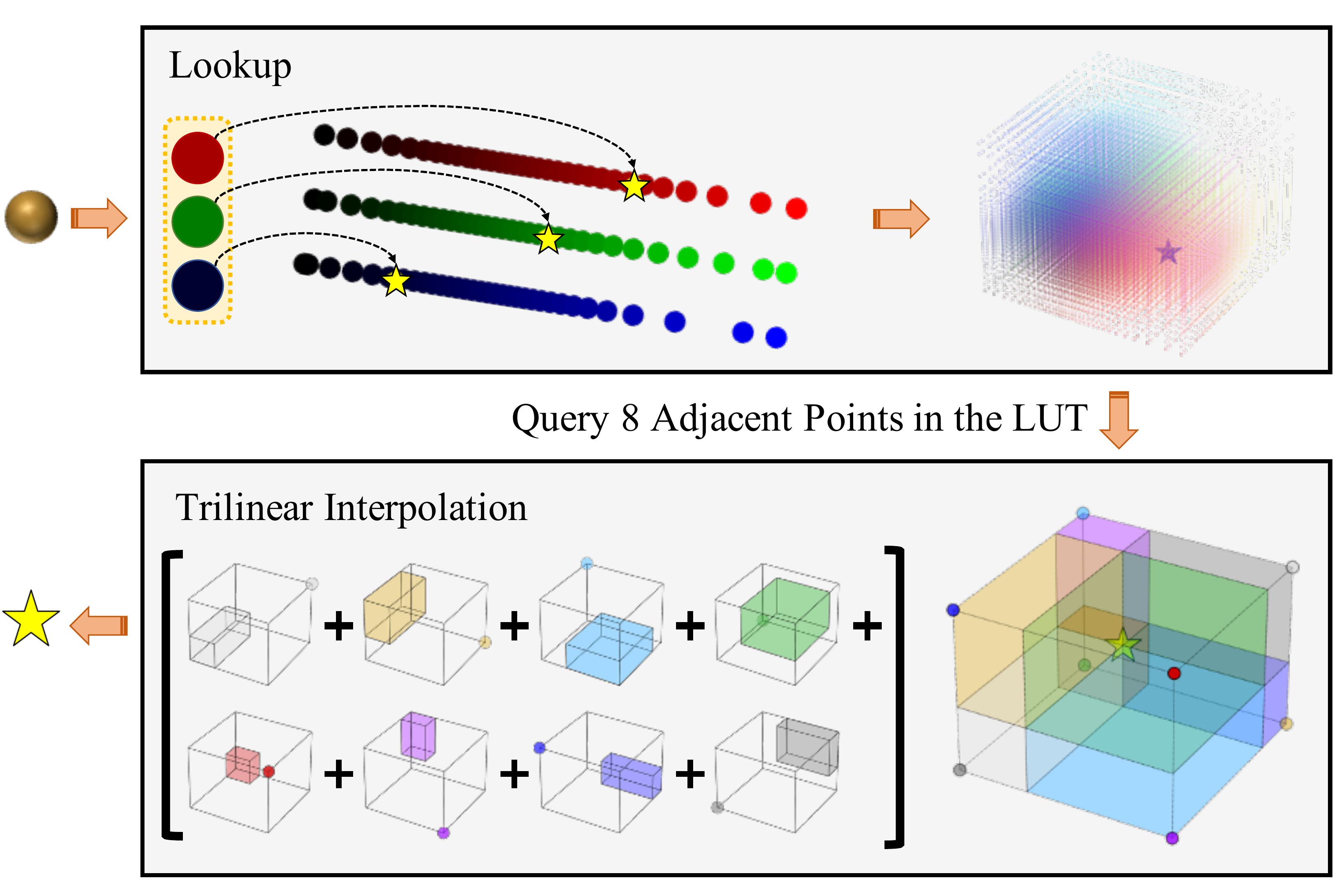}
  \caption{Procedure of the proposed \textit{AiLUT-Transform}, which is achieved by two operations: lookup and interpolation. Best viewed in color.}
  \label{fig:ailut-transform}
  \vspace{-0.4cm}
\end{figure}

\subsection{Differentiable Adaptive Interval Lookup Table Transform (AiLUT-Transform)}
\label{sec:method-ailut-transform}

With the involvement of AdaInt, the LUT transform should take both the output values $T$ and the sampling coordinates $P$ of the LUT, along with the input image $X$ to produce the transformed output image $\hat{Y}$. In the standard LUT transform, $P$ is usually omitted since the sampling coordinates are assumed uniform. Therefore, the gradient with respect to $P$ has not yet been explored, which hinders the end-to-end learning of AdaInt. To this end, we introduce a novel transform operation called \textit{AiLUT-Transform}:
\begin{equation}
    \hat{Y} = \text{AiLUT-Transform}(X, T, P).
    \label{eq:ailut-transform}
\end{equation}
The \textit{AiLUT-Transform} is (sub-)differential with respect to not only $X$ and $T$, but also $P$. This enables the end-to-end learning of the AdaInt module. Given an input query pixel $x$ consisting of three color components $\{x_r, x_g, x_b\}$, \textit{AiLUT-Transform} computes its transformed color via two basic steps: lookup and interpolation. Please also see \Cref{fig:ailut-transform} for a graphic illustration.

\vspace{-0.2cm}

\paragraph{The Lookup Step}
Our \textit{AiLUT-Transform} first performs a lookup operation to locate the query pixel in the 3D LUT. As shown in the top part of \Cref{fig:ailut-transform}, this operation aims to find both the left and right neighbors $x_c^0, x_c^1\in P\ (c=r,g,b)$ along each dimension for the query pixel. It can be easily achieved by a \textit{binary search} thanks to the bounded and the monotone increasing properties of our learned sampling coordinates (see \Cref{sec:method-adaint}). Accordingly, the $8$ adjacent points in the LUT can be queried using the indices of the located neighbors in $P$. For a sampling point corresponding to $x_r^i, x_g^j, x_b^k$, where $i,j,k\in\{0,1\}$, we abbreviate the output color values of these 8 neighbors as $\tilde{T}_{:,i,j,k}$.

\vspace{-0.2cm}

\paragraph{The Interpolation Step}
After querying 8 adjacent points, trilinear interpolation is conducted to compute the transformed output color of the query pixel. As shown in the bottom part of \Cref{fig:ailut-transform}, the transformed output $\hat{y}$ is the sum of the values at $8$ corners weighted by the \textit{normalized} partial volume diagonally opposite the corners, which can be formulated as:
\begin{equation}
\begin{split}
    \hat{y} = \sum_{i,j,k\in\{0,1\}}V_{i,j,k}\cdot\tilde{T}_{:,i,j,k},
\end{split}
\end{equation}
where $V_{i,j,k} = (x_r^d)^i(1-x_r^d)^{1-i}(x_g^d)^j(1-x_g^d)^{1-j}(x_b^d)^k(1-x_b^d)^{1-k}$, and $x_c^d = (x_c - x_c^0) / (x_c^1 - x_c^0)\ (c=r,g,b)$.

\vspace{-0.2cm}

\paragraph{Backpropagation}
To allow the learning of AdaInt via backpropagation, we derive the gradients with respect to $x_c^0, x_c^1$, and therefore to $P$. The partial derivative of $x_c^0$ is:
\begin{equation}
\label{eq:adic-gradient}
    \frac{\partial \hat{y}}{\partial x_c^0} = \sum_{i,j,k\in\{0,1\}}\tilde{T}_{:,i,j,k}\frac{\partial V_{i,j,k}}{\partial x_c^d}\frac{\partial x_c^d}{\partial x_c^0}
\end{equation}
and similarly to \Cref{eq:adic-gradient} for $x_c^1$. Please refer to \iftoggle{arxiv}{\Cref{sec:suppl-backprop}}{the supplementary material} for detailed derivation. Besides, the gradient with respect to $\tilde{T}_{:,i,j,k}$ is more concise: $\partial \hat{y}/\partial \tilde{T}_{:,i,j,k} = V_{i,j,k}$.

As the proposed \textit{AiLUT-Transform} is applied to each pixel independently, it can be implemented efficiently via CUDA. We merge the lookup and interpolation operations into a single CUDA kernel to maximize parallelism. Since our lookup operation is achieved by the binary search algorithm of logarithmic time complexity ($\mathcal{O}(\log_2 N_s)$), its computational cost is negligible in our case, where $N_s$ has a relatively small value (typically, 33).

\begin{figure}
    \centering
    \includegraphics[width=\linewidth]{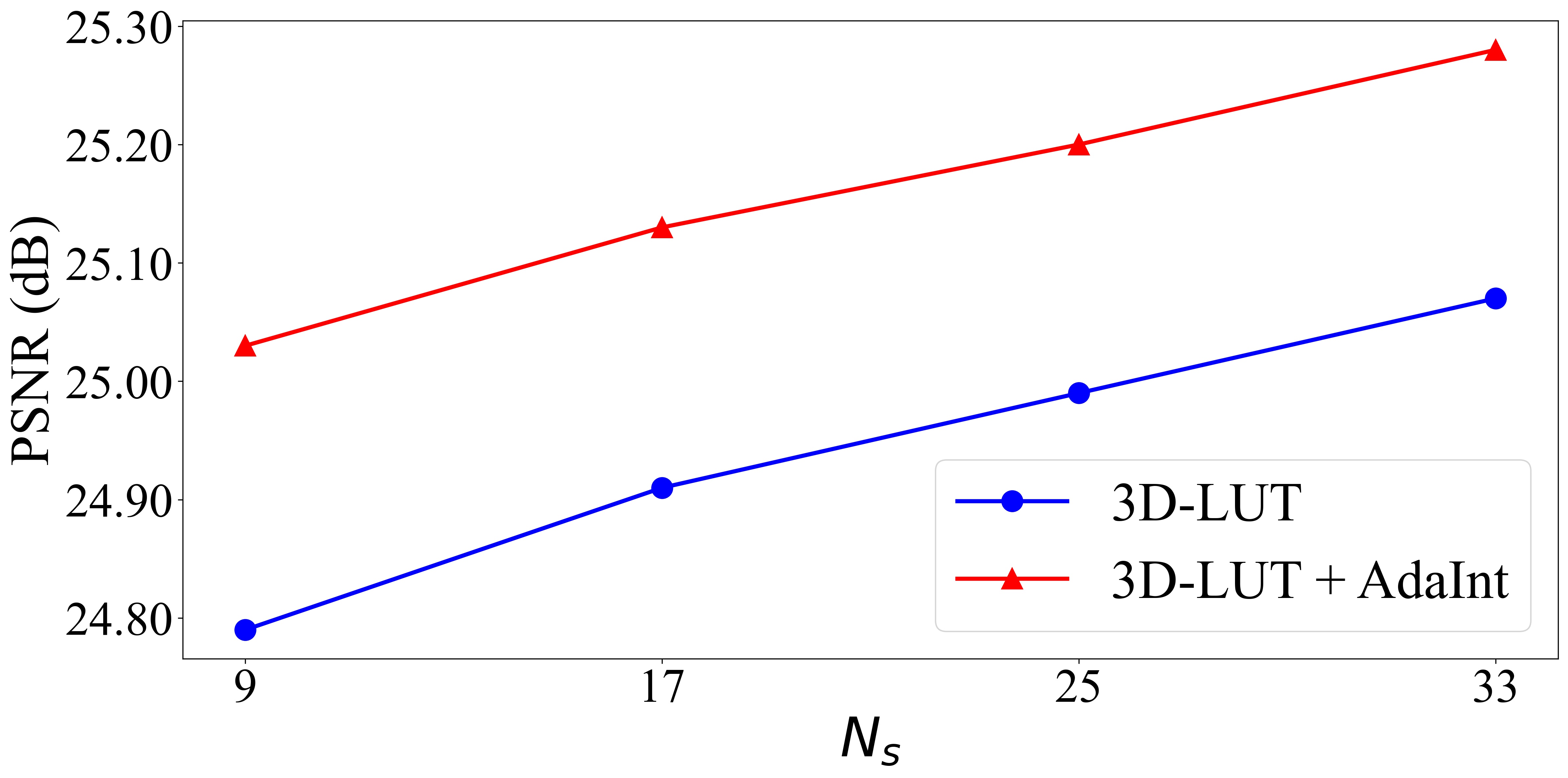}
    \caption{Ablation study on AdaInt under different numbers ($N_s$) of sampling coordinates. The results on the \textbf{FiveK} dataset (480p)~\cite{CVPR11FiveK} for \textbf{tone mapping} are plotted.}
    \label{fig:abl-n-adic}
\end{figure}

\begin{table}[t]
    \centering
    \begin{tabular}{ccc}
        \toprule[1pt]
        Sampling Strategy & PSNR$\uparrow$ & SSIM$\uparrow$ \\
        \midrule[1pt]
        Shared-AdaInt & 25.13 & 0.921 \\
        AdaInt & 25.28 & 0.925 \\
        \bottomrule[1pt]
    \end{tabular}
    \caption{Ablation study on different sampling strategies in AdaInt. The results on the \textbf{FiveK} dataset (480p)~\cite{CVPR11FiveK} for \textbf{tone mapping} are listed. "$\uparrow$" indicates the larger is better.}
    \label{tab:abl-share-adic}
    \vspace{-0.4cm}
\end{table}

\begin{table*}[ht]
    \centering
    \setlength{\tabcolsep}{7pt}
    \begin{tabular}{cccccccccc}
    \toprule[1pt]
    \multirow{2}{*}{Method} &
    \multirow{2}{*}{\#Parameters} & \multicolumn{4}{c}{480p} & \multicolumn{4}{c}{Full Resolution (4K)} \\
    \cline{3-10}
    & & PSNR & SSIM & $\Delta E_{ab}$ & Runtime & PSNR & SSIM & $\Delta E_{ab}$ & Runtime \\
    \midrule[1pt]
    UPE~\cite{CVPR19UPE} & 927.1K & 21.88 & 0.853 & 10.80 & 4.27 & 21.65 & 0.859 & 11.09 & 56.88 \\
    DPE~\cite{CVPR18DPE} & 3.4M & 23.75 & 0.908 & 9.34 & 7.21 & - & - & - & - \\
    HDRNet~\cite{TOG17HDRNet} & 483.1K & 24.66 & 0.915 & 8.06 & 3.49 & 24.52 & 0.921 & 8.20 & 56.07 \\
    DeepLPF~\cite{CVPR20DeepLPF} & 1.7M & 24.73 & 0.916 & 7.99 & 32.12 & - & - & - & - \\
    CSRNet~\cite{ECCV20CSRNet} & 36.4K & 25.17 & \textcolor{blue}{0.924} & 7.75 & 3.09 & 24.82 & 0.926 & 7.94 & 77.10 \\
    SA-3DLUT~\cite{ICCV21SALUT}* & 4.5M & \textcolor{red}{25.50} & / & / & 2.27 & / & / & / & 4.39 \\
    \midrule[1pt]
    3D-LUT~\cite{TPAMI203DLUT} & 593.5K & 25.29 & 0.923 & \textcolor{blue}{7.55} & \textcolor{red}{1.17} & \textcolor{blue}{25.25} & \textcolor{blue}{0.932} & \textcolor{blue}{7.59} & \textcolor{red}{1.49} \\
    3D-LUT + AdaInt & 619.7K & \textcolor{blue}{25.49} & \textcolor{red}{0.926} & \textcolor{red}{7.47} & \textcolor{blue}{1.29} & \textcolor{red}{25.48} & \textcolor{red}{0.934} & \textcolor{red}{7.45} & \textcolor{blue}{1.59} \\
    \bottomrule[1pt]
    \end{tabular}
    \caption{Quantitative comparisons on the \textbf{FiveK} dataset ~\cite{CVPR11FiveK} for \textbf{photo retouching}. Runtime is measured in milliseconds. "-" means the result is not available due to insufficient GPU memory. The "*" symbol indicates that the results are adopted from the original paper (some are absent ("/")) due to the unavailable source code. The best and second results are highlighted in \textcolor{red}{red} and \textcolor{blue}{blue}, respectively.}
    \label{tab:sota-fivek-srgb}
    \vspace{-0.4cm}
\end{table*}

\subsection{Loss Function}
\label{sec:method-loss}

The overall framework can be trained in an end-to-end manner. Our loss function consists of the MSE loss as the reconstruction loss ($\mathcal{L}_r$) and some regularization terms adopted from ~\cite{TPAMI203DLUT} to constrain the output values $T$ of the LUT, including smoothness term ($\mathcal{L}_s$) and monotonicity term ($\mathcal{L}_m$). We do not introduce any other constraint or loss function to the learning of AdaInt, willing that it can be \textit{image-adaptive} for the network. Following~\cite{TPAMI203DLUT}, our final loss is written as:
\begin{equation}
    \label{eq:loss}
    \mathcal{L} = \mathcal{L}_r + 0.0001\times\mathcal{L}_s + 10\times\mathcal{L}_m.
\end{equation}
\section{Experiments}
\label{sec:experiments}

\begin{table}[ht]
    \centering
    \begin{tabular}{cccc}
    \toprule[1pt]
    \multirow{2}{*}{Method} & \multicolumn{3}{c}{480p} \\
    \cline{2-4}
                            & PSNR   & SSIM   & $\Delta E_{ab}$  \\
    \midrule[1pt]
    UPE~\cite{CVPR19UPE} & 21.56 & 0.837 & 12.29 \\
    DPE~\cite{CVPR18DPE} & 22.93 & 0.894 & 11.09 \\
    HDRNet~\cite{TOG17HDRNet} & 24.52 & 0.915 & 8.14 \\
    CSRNet~\cite{ECCV20CSRNet} & \textcolor{blue}{25.19} & \textcolor{blue}{0.921} & 7.63 \\
    \midrule[1pt]
    3D-LUT~\cite{TPAMI203DLUT} & 25.07 & 0.920 & \textcolor{blue}{7.55} \\
    3D-LUT + AdaInt         & \textcolor{red}{25.28} & \textcolor{red}{0.925} & \textcolor{red}{7.48} \\
    \bottomrule[1pt]
    \end{tabular}
    \caption{Quantitative comparisons on the \textbf{FiveK} dataset (480p)~\cite{CVPR11FiveK} for the \textbf{tone mapping} application. The best and second results are highlighted in \textcolor{red}{red} and \textcolor{blue}{blue}, respectively.}
    \label{tab:sota-fivek-xyz}
    \vspace{-0.4cm}
\end{table}

\subsection{Datasets and Application Settings}
\label{sec:exp-datasets}

We evaluate our method on two publicly available datasets: MIT-Adobe FiveK~\cite{CVPR11FiveK} and PPR10K~\cite{ICCV21PPR10K}. The MIT-Adobe FiveK is a commonly used photo retouching dataset with 5,000 RAW images. We follow the common practice in recent works~\cite{TPAMI203DLUT, ECCV20CSRNet, ICCV21RCT} to adopt only the version retouched by expert C as the groundtruth and split the dataset into 4,500 image pairs for training and 500 image pairs for testing. To speed up the training stage, images are downsampled to 480p resolution (with the short side resized to 480 pixels), whereas images of both 480p and original 4K resolutions are used during testing. The PPR10K is a newly released portrait photo retouching dataset with a larger scale of 11,161 high-quality RAW portrait photos. All three retouched versions are used as the groundtruth in three separable experiments. Following the official split~\cite{ICCV21PPR10K}, we divide the dataset into 8,875 pairs for training and 2,286 pairs for testing. Experiments are conducted on the 360p version of the dataset due to insufficient disk space. Please refer to \iftoggle{arxiv}{\Cref{sec:appl-impl}}
{the supplementary materials} for more details.

We follow ~\cite{TPAMI203DLUT} to conduct our experiments on two typical applications: \textit{photo retouching} and \textit{tone mapping}. The target images in both applications share the same 8-bit sRGB format. The difference between the two tasks lies in the input formats. In the photo retouching task, the input images are also in sRGB format (8-bit on FiveK and 16-bit on PPR10K), while for the tone mapping task, the input images are in 16-bit CIE XYZ format. Therefore, the tone mapping task requires the ability of color space conversion. We conduct both tasks on the FiveK dataset, but only the retouching task on PPR10K as done in~\cite{ICCV21PPR10K}.

\begin{table}[ht]
    \centering
    \begin{tabular}{P{1.20in}P{0.03in}P{0.25in}P{0.25in}P{0.3in}P{0.3in}}
        \toprule[1pt]
        {\small Method} & {\small E} & {\small PSNR} & {\small $\Delta E_{ab}$} & {\small $\text{PSNR}^\text{HC}$} & {\small $\Delta E_{ab}^\text{HC}$} \\
        \midrule[1pt]
        HDRNet~\cite{TOG17HDRNet}  & a & 23.93 & 8.70 & 27.21 & 5.65 \\
        CSRNet~\cite{ECCV20CSRNet} & a & 22.72 & 9.75 & 25.90 & 6.33 \\
        3D-LUT~\cite{TPAMI203DLUT} & a & 25.64 & 6.97 & \textcolor{blue}{28.89} & 4.53 \\
        3D-LUT + HRP~\cite{ICCV21PPR10K} & a & \textcolor{blue}{25.99} & \textcolor{blue}{6.76} & 28.29 & \textcolor{blue}{4.38} \\
        3D-LUT + AdaInt & a & \textcolor{red}{26.33} & \textcolor{red}{6.56} & \textcolor{red}{29.57} & \textcolor{red}{4.26} \\
        \midrule[1pt]
        HDRNet~\cite{TOG17HDRNet}  & b & 23.96 & 8.84 & 27.21 & 5.74 \\
        CSRNet~\cite{ECCV20CSRNet} & b & 23.76 & 8.77 & 27.01 & 5.68 \\
        3D-LUT~\cite{TPAMI203DLUT} & b & 24.70 & 7.71 & 27.99 & 4.99 \\
        3D-LUT + HRP~\cite{ICCV21PPR10K} & b & \textcolor{blue}{25.06} & \textcolor{blue}{7.51} & \textcolor{blue}{28.36} & \textcolor{blue}{4.85} \\
        3D-LUT + AdaInt & b & \textcolor{red}{25.40} & \textcolor{red}{7.33} & \textcolor{red}{28.65} & \textcolor{red}{4.75} \\
        \midrule[1pt]
        HDRNet~\cite{TOG17HDRNet}  & c & 24.08 & 8.87 & 27.32 & 5.76 \\
        CSRNet~\cite{ECCV20CSRNet} & c & 23.17 & 9.45 & 26.47 & 6.12 \\
        3D-LUT~\cite{TPAMI203DLUT} & c & 25.18 & 7.58 & 28.49 & 4.92 \\
        3D-LUT + HRP~\cite{ICCV21PPR10K} & c & \textcolor{blue}{25.46} & \textcolor{blue}{7.43} & \textcolor{blue}{28.80} & \textcolor{blue}{4.82} \\
        3D-LUT + AdaInt & c & \textcolor{red}{25.68} & \textcolor{red}{7.31} & \textcolor{red}{28.93} & \textcolor{red}{4.76} \\
        \bottomrule[1pt]
    \end{tabular}
    \caption{Quantitative comparisons on the \textbf{PPR10K} dataset \cite{ICCV21PPR10K} for \textbf{portrait photo retouching}, where "E" denotes "Expert", and a, b, c indicate the groundtruths retouched by three experts.}
    \label{tab:sota-ppr10k}
    \vspace{-0.4cm}
\end{table}

\begin{figure*}[t]
    \centering
    \includegraphics[width=\linewidth]{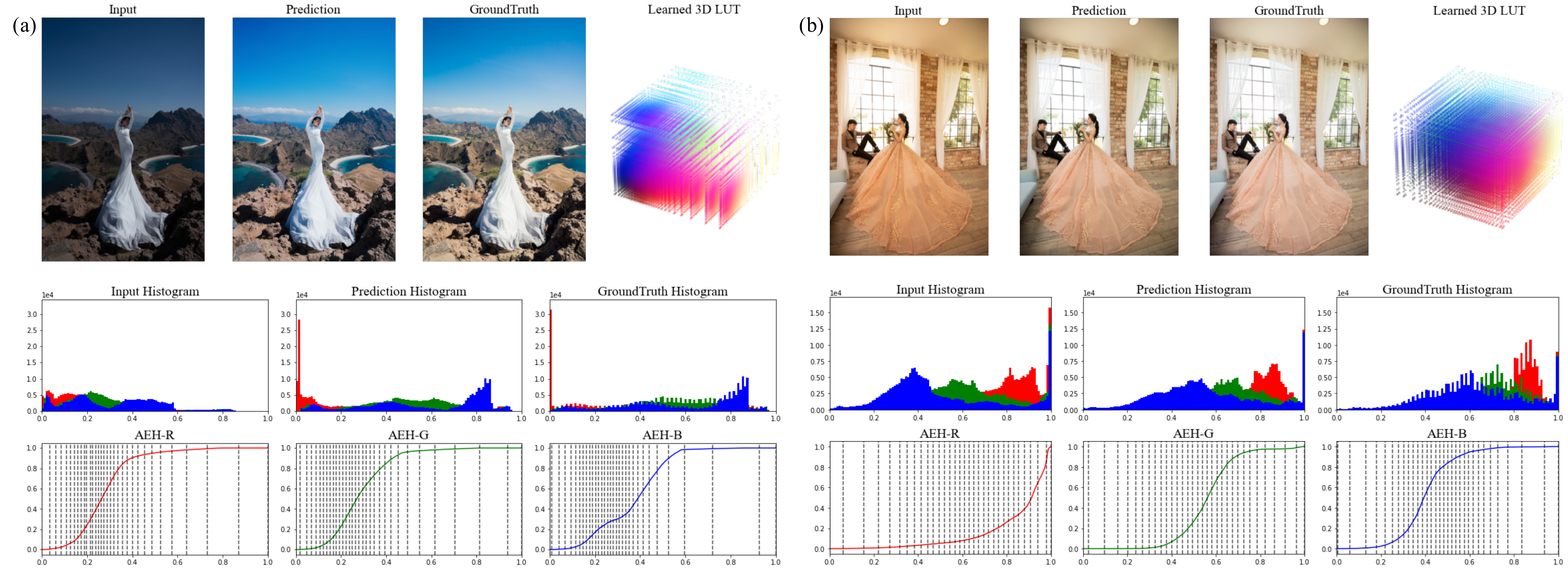}
    \caption{Illustration of the learned sampling coordinates and the corresponding 3D LUTs for \textbf{photo retouching} on the \textbf{PPR10K} dataset (360p)~\cite{ICCV21PPR10K}. The bottom row visualizes the learned sampling coordinates on the so-called per-color-channel \textit{Accumulative Error Histogram (AEH)}~\cite{ECCV12Nonuniform}. The regions in the AEH exhibiting high curvature indicate wherein more sampling points are needed. Best viewed on screen.}
    \label{fig:learned-intervals}
    \vspace{-0.4cm}
\end{figure*}

\subsection{Implementation Details}
\label{sec:exp-impl}

Since the focus of our work is to present the idea of learning image-adaptive sampling intervals for a 3D LUT, we do not dive into complicated architectural engineering. Instead, to instantiate the mapping $f$ in our method, we directly follow Zeng's~\cite{TPAMI203DLUT, ICCV21PPR10K} practices to adopt the 5-layer backbone in ~\cite{TPAMI203DLUT} on the FiveK dataset and the ResNet-18~\cite{ResNet} (initialized with ImageNet-pretrained~\cite{ImageNet} weights) on the PPR10K dataset. The mapping $h$ in \Cref{eq:outval} is implemented with two cascade fully-connected layers, which in practice reformulates the implementation in \cite{TPAMI203DLUT}. For the instantiation of AdaInt (mapping $g$ in \Cref{eq:adaint}), a single fully-connected layer is employed. The weights and bias of $g$ are initialized to 0s and 1s, which makes the predicted sampling intervals start from a uniform state. Please refer to \iftoggle{arxiv}{\Cref{sec:appl-impl}}{the supplementary materials} for more details.

We use the standard Adam optimizer~\cite{Adam} to minimize the loss function in \Cref{eq:loss}. The mini-batch size is set to 1 and 16 on FiveK and PPR10K, respectively. All our models are trained for 400 epochs with a fixed learning rate of $1\times 10^{-4}$. We decay the learning rate of $g$ by a factor of 0.1 and freeze its parameters in the first 5 training epochs to make the AdaInt learning more stable. Our method is implemented based on PyTorch~\cite{PyTorch}. All experiments are conducted on an NVIDIA Tesla V100 GPU. The settings of $N_s$ and $M$ are according to the datasets and the experimental purposes. We provide them in the following sections. 

\subsection{Ablation Studies}
\label{sec:exp-abla}

In this section, the tone mapping task with images from the FiveK dataset (480p) is chosen to conduct several ablation studies for verifying the proposed AdaInt. We expect the higher dynamic range (16-bit) of the input images in the tone mapping task can better examine the ability of our AdaInt to learn image-adaptive sampling points. In all ablation studies, the hyper-parameter $M$ is set to 3.

\vspace{-0.4cm}

\paragraph{Number of Coordinates along Each Dimension}
We assess the baseline 3D-LUT~\cite{TPAMI203DLUT} and our method under different settings of $N_s$ (the number of sampling coordinates along each color dimension) to verify the efficacy of the proposed AdaInt. As shown in \Cref{fig:abl-n-adic}, the performance of the baseline and our method decrease as a smaller $N_s$ is adopted. Our AdaInt consistently improves the baseline under all settings $N_s$.
Further increasing $N_s$ (from 33 to 65) can only bring marginal improvement (0.05dB) on the baseline compared with that introduced by our AdaInt. 
It is worth noting that our method achieves comparable or even better performance with a relatively small LUT size ($N_s$) compared to the baseline. It is because AdaInt enables the ability of 3D LUTs to take full advantage of the limited sampling points for better modeling on the underlying optimal color transform.

\vspace{-0.4cm}

\paragraph{Sampling Strategy}
Our AdaInt generates an individual set of sampling intervals for each color dimension separably, making our method adopts different sampling strategies along different color dimensions. It divides the entire 3D color space into various cuboids. Here, we compare such a default setting with another one that adopts the same strategy over three color dimensions, which divides the 3D space into cubes. We achieve it by letting AdaInt generate only a set of sampling intervals and replicate it to three color dimensions, abbreviated as \textit{Shared-AdaInt}. As shown in \Cref{tab:abl-share-adic}, the Shared-AdaInt strategy performs inferior to the default setting, which is in line with our expectation as the sharing mechanism limits the flexibility of AdaInt to allocate sampling points in the 3D space.

\begin{figure*}[ht]
    \centering
    \includegraphics[width=\linewidth]{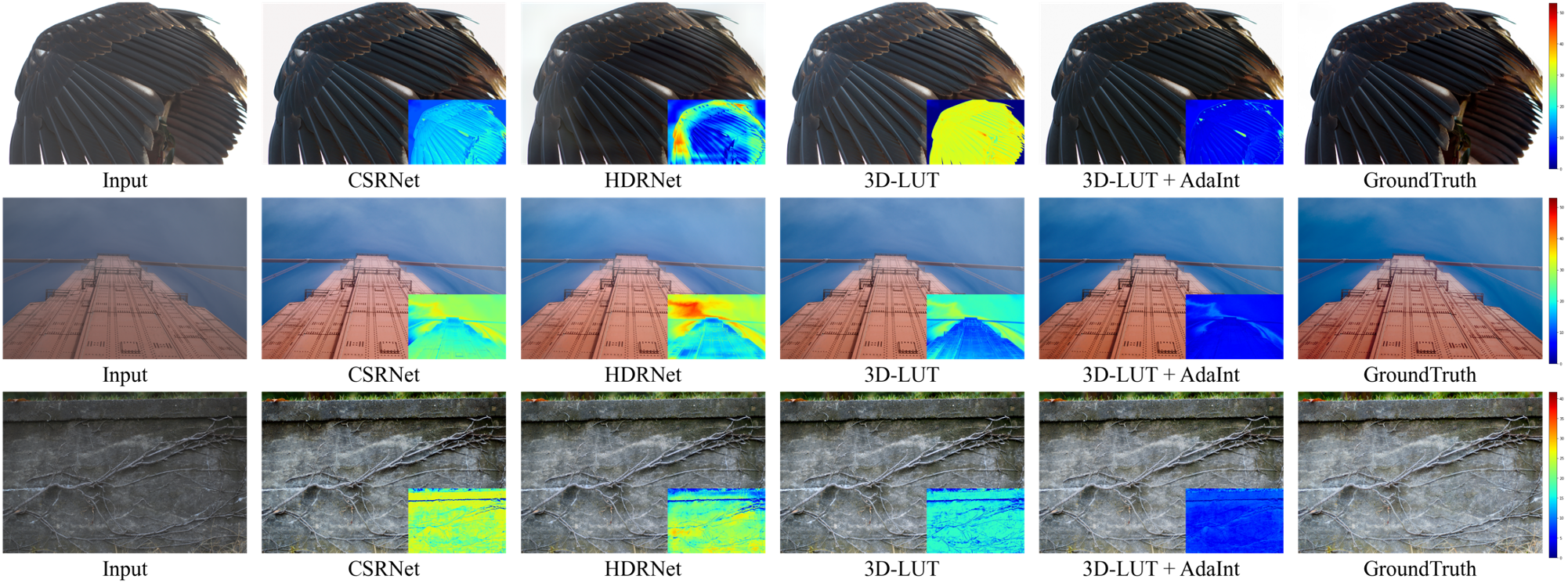}
    \caption{Qualitative comparisons with corresponding error maps on the \textbf{FiveK} dataset \cite{CVPR11FiveK} for \textbf{photo retouching}. Best viewed on screen.}
    \label{fig:sota-quality}
    \vspace{-0.4cm}
\end{figure*}

\subsection{Property of the Adaptive Sampling Intervals}
\label{sec:exp-analysis}

The top part of \Cref{fig:learned-intervals} shows two different photos on the PPR10K dataset, their color histograms, and the corresponding learned 3D LUTs from our model. It can be observed that both the color and layout of the 3D lattices vary with the different image content, indicating the image-adaptive property of our learned 3D LUTs. To better analyze the behavior of our AdaInt, we introduce the per-color-channel \textit{Accumulative Error Histogram (AEH)}~\cite{ECCV12Nonuniform} between the input and groundtruth images. The regions with high curvature in the AEH, to some extent, indicate the complexity/local-nonlinearity of the underlying 3D color transform and hence require more sampling points. As shown in the bottom part of \Cref{fig:learned-intervals}, the sampling coordinates predicted by our AdaInt non-uniformly and adaptively distribute to different regions according to the transform complexity on various images and color channels. A detailed description of AEH and more visualization of learned intervals can be found in \iftoggle{arxiv}{\Cref{sec:appl-adaint}}{the supplementary materials}.

\subsection{Comparison with State-of-the-Arts}
\label{sec:exp-sota}

We also compare \textit{state-of-the-art} \textit{real-time} photo enhancement methods. $N_s$ is set to 33 as done in other 3D LUT-based approaches~\cite{TPAMI203DLUT, ICCV21SALUT} for fair comparisons. $M$ is set to 3 and 5 for the FiveK and PPR10K datasets, respectively, as done in~\cite{ICCV21PPR10K}.

\vspace{-0.4cm}

\paragraph{Quantitative Comparisons}
We compare the selected methods on PSNR, SSIM~\cite{
SSIM}, the $L_2$-distance in CIE LAB color space ($\Delta E_{ab}$), and the inference speed. On PPR10K, we also include the human-centered measures~\cite{ICCV21PPR10K} (denoted by the "HC" superscript). 
We obtain the results of existing methods using their published codes and default configurations. All approaches are executed on an NVIDIA Tesla V100 GPU. For speed comparison, we 
measure the GPU inference time on 100 images and report the average.
\Cref{tab:sota-fivek-srgb} lists the comparison on the FiveK for photo retouching.
Our method outperforms others with relatively fewer parameters on both resolutions. Similar conclusions apply to \Cref{tab:sota-fivek-xyz,tab:sota-ppr10k} on the FiveK for tone mapping and the PPR10K for portrait photo retouching, respectively. Especially, our AdaInt brings consistent improvement over 3D-LUT~\cite{TPAMI203DLUT} on all datasets with a negligible computational cost increase, demonstrating its efficiency and effectiveness. It is worth noting that the concurrent study SA-3DLUT~\cite{ICCV21SALUT} promotes 3D LUTs by constructing pixel-wise LUTs at the cost of a significant model size increase (about 7 times) and a speed decrease (about 3 times). We believe SA-3DLUT equipped with our AdaInt can be further improved, though the source code is not yet publicly available.

\vspace{-0.4cm}

\paragraph{Qualitative Comparisons}
\Cref{fig:sota-quality} shows that our method produces more visually pleasing results than other methods. For example, our method better handles the overexposure of the image in the first row.
In the second row, other methods suffer from poor saturation in the blue sky, resulting in hazy photos. Our AdaInt instead successfully produces the correct blue color and thus provides a cleaner result. Besides, when enhancing the brightness in the third row, our method preserves more rock texture. Please refer to \iftoggle{arxiv}{\Cref{fig:sota-fivek-1,fig:sota-ppr10k-1}}{the supplementary materials} for more comparisons.
\section{Limitation and Conclusion}
\label{sec:conclusion}

While our AdaInt promotes the expressiveness of 3D LUTs by providing image-adaptive sampling strategies, it still lacks spatial modeling and noise robustness. The 3D LUTs assume that each pixel is transformed independently according to its color without considering the locality. Hence, it is more suited for global enhancement and may produce less satisfactory results in areas requiring local tone mapping. ~\cite{ICCV21SALUT} provided a possible solution by constructing pixel-wise LUTs. Our method is orthogonal to and may also bring improvement over it. Besides, as our approach is based on pixel-wise mapping, heavy noise may also influence our results. Please refer to \iftoggle{arxiv}{\Cref{fig:noise}}{the supplementary materials} for some visual examples.

In this paper, we present AdaInt, a novel learning mechanism to promote learnable 3D LUTs for real-time image enhancement. The central idea is to introduce image-adaptive sampling intervals for learning a non-uniform 3D LUT layout. We develop AdaInt as a plug-and-play neural network module and propose a differentiable AiLUT-Transform operator encapsulating binary search and trilinear interpolation. Experimental results on two datasets demonstrate the superiority of our method over other \textit{state-of-the-art} methods in terms of both performance and efficiency.
In addition, we believe the viewpoint of non-uniform sampling on a complicated underlying transform function or representation is not limited to 3D LUTs and can also facilitate other applications, which we leave as our future work.

\vspace{-0.1cm}
\vspace*{\fill}

\section*{Acknowledgement}

Yi Xu is supported in part by the National Natural Science Foundation of China (62171282, 111 project BP0719010, STCSM 18DZ2270700), the Shanghai Municipal Science and Technology Major Project (2021SHZDZX0102), and the Key Research and Development Program of Chongqing (cstc2021jscx-gksbX0032). Canqian Yang is supported in part by Alibaba Group through Alibaba Research Intern Program.

\clearpage

{\small
\bibliographystyle{ieee_fullname}
\bibliography{egbib}
}
\clearpage

\appendix

\section{Motivation of Using 3D LUT}

Inspired by the practice in the image signal processor (ISP) \cite{ECCV16software}, we adopt the 3D LUT instead of a more general mapping function (\textit{e.g.}, a multi-layer perceptron (MLP)) to model the $\mathbb{R}^3\rightarrow\mathbb{R}^3$ color transform function due to the consideration of \textit{efficiency and expressiveness}. Specifically, if a \textit{static} MLP is used, it is easy to learn an over smooth transform since the network needs to adapt to all images in the dataset. Therefore, the model might suffer from its limited expressiveness and diversity among images. Adopting an additional network for predicting the parameters of an MLP to enable image-adaptiveness is another alternative and has been investigated in CSRNet \cite{ECCV20CSRNet}. However, since an MLP requires a cascade of several linear and nonlinear sub-operations to increase the model capability, it suffers from a higher computational burden, especially on high-resolution inputs, as shown in \iftoggle{selfappendix}{Table 2 in the paper}{\Cref{tab:sota-fivek-srgb}} (\textbf{48 times} over our AdaInt on 4K resolution). Instead, our 3D LUT with learned adaptive intervals is able to achieve enhanced expressiveness and image-adaptiveness while still presenting high efficiency by directly recording and retrieving a complex transform via simple lookup and interpolation operations.

\section{Details of AiLUT-Transform}
\label{sec:appd-ailut-transform}

\subsection{Forward}

Given an input image denoted as $X\in[0,1]^{3\times H\times W}$, suppose an image-adaptive 3D LUT is learned, its output values and sampling coordinates are abbreviated as $T\in[0,1]^{3\times N_s\times N_s\times N_s}$ and $\hat{P}\in[0,1]^{3\times N_s}$, respectively. $N_s$ is the number of sampling coordinates along each lattice dimension. The AiLUT-Transform takes all of $X$, $T$, and $\hat{P}$ as inputs, and produces the transformed output image $\hat{Y}$:
\begin{equation}
    \hat{Y} = \text{AiLUT-Transform}(X,T,\hat{P}).
\end{equation}
The transform consists of a lookup step and an interpolation step. The former locates each input pixel of $X$ in the LUT, whereas the latter computes the corresponding output using the values of eight nearest neighborhood sampling points in $T$. Suppose an input query pixel $x$ is given, which is composed of three color components $\{x_r,x_g,x_b\}$, the transform on $x$ is performed in the following steps.

\paragraph{The Lookup Step}
The transform first locates the query pixel $x$ into a lattice cell closed by 8 nearest neighborhood sampling points. These 8 points (or called vertices) are determined by 6 coordinates obtained by finding both the left and right neighbors $x_c^0,x_c^1\in \hat{P}\ (c=r,g,b)$ along each lattice dimension, and their corresponding indices in the LUT $e_c^0,e_c^1\in\mathbb{I}_0^{N_s-1}$. In a standard LUT transform where the 3D lattice is constructed with equal intervals $\Delta=1/(N_s-1)$, this can be easily achieved:
\begin{align*}
    e_c^0=\lfloor\frac{x_c}{\Delta}\rfloor, e_c^1=e_c^0+1,\\
    x_c^0=e_c^0\Delta, x_c^1=e_c^1\Delta,
\end{align*}
where $\lfloor\cdot\rfloor$ is the floor function. It is obvious that the $\hat{P}$ is not needed in the standard LUT transform, so it is always omitted in existing implementations. However, in AiLUT-Transform, since the intervals are no longer a constant value, a searching algorithm on $\hat{P}$ is required to find $x_c^0,x_c^1$ and $e_c^0,e_c^1$, which can be formulated as:
\begin{equation}
\label{eq:binary-search}
\begin{split}
    x_c^0, e_c^0 &= \max \{\hat{P}_{c,s}, s|\hat{P}_{c,s}\leq x_c, s\in\mathbb{I}_0^{N_s-1}\}, \\
    x_c^1, e_c^1 &= \min \{\hat{P}_{c,s}, s|\hat{P}_{c,s}> x_c, s\in\mathbb{I}_0^{N_s-1}\}.
\end{split}
\end{equation}
Thanks to the bounded and monotone increasing properties of the learned sampling coordinates stored in $\hat{P}$ (see \iftoggle{selfappendix}{Section 3.2 in the paper}{\Cref{sec:method-adaint}}), \Cref{eq:binary-search} can be easily achieved by introducing a binary search. 

Once the 6 indices ($e_r^0,e_r^1;e_g^0,e_g^1;e_b^0,e_b^1$) for the nearest neighborhood vertices in the LUT are determined, the corresponding output color values of the 8 neighbors in $T$ can be defined as:
\begin{equation}
    \tilde{T}_{:,i,j,k}=T[:, e_r^i, e_g^j, e_b^k],\ i,j,k\in\{0,1\}.
\end{equation}

\paragraph{The Interpolation Step}
After querying 8 adjacent vertices in the LUT for query pixel $x$, the transformed result can be obtained by applying trilinear interpolation, including the following steps:
\begin{itemize}
    \item Compute the normalized offsets between the input color and the 8 neighbors:
    \begin{equation}
        x_r^d=\frac{x_r-x_r^0}{x_r^1-x_r^0}, x_g^d=\frac{x_g-x_g^0}{x_g^1-x_g^0}, x_b^d=\frac{x_b-x_b^0}{x_b^1-x_b^0}.
    \end{equation}
    \item Compute the interpolation weights, which are defined as the partial volume diagonally opposite the vertices:
    \begin{align}
    \begin{split}
        V_{i,j,k}=&(x_r^d)^i(1-x_r^d)^{1-i}\\
        &\cdot(x_g^d)^j(1-x_g^d)^{1-j}\\
        &\cdot(x_b^d)^k(1-x_b^d)^{1-k}.
    \end{split}
    \end{align}
    \item Compute the linear combination of $V_{i,j,k}$ and $\tilde{T}_{:,i,j,k}$:
    \begin{equation}
        \hat{y}=\sum_{i,j,k\in\{0,1\}}V_{i,j,k}\cdot\tilde{T}_{:,i,j,k}.
    \end{equation}
\end{itemize}

\subsection{Backpropagation}
\label{sec:suppl-backprop}

In our method, both $T$ and $\hat{P}$ are learned by neural networks to enable more flexible 3D LUT with adaption to different image content. The challenge exists in the absence of interpolation implementation in popular deep learning libraries (such as PyTorch \cite{PyTorch}) that can provide gradients to $\hat{P}$. To this end, we provide a new implementation that can compute the gradients with respect to $\hat{P}$ during backpropagation to enable the end-to-end learning of AdaInt. The core is to define the partial derivatives of $x_c^0,x_c^1$:
\begin{equation}
\label{eq:backward-1}
\begin{split}
    \frac{\partial \hat{y}}{\partial x_c^0} = \sum_{i,j,k\in\{0,1\}}\tilde{T}_{:,i,j,k}\frac{\partial V_{i,j,k}}{\partial x_c^d}\frac{\partial x_c^d}{\partial x_c^0},\\
    \frac{\partial \hat{y}}{\partial x_c^1} = \sum_{i,j,k\in\{0,1\}}\tilde{T}_{:,i,j,k}\frac{\partial V_{i,j,k}}{\partial x_c^d}\frac{\partial x_c^d}{\partial x_c^1}.
\end{split}
\end{equation}
where,
\begin{equation}
\label{eq:backward-2}
\begin{split}
    \frac{\partial V_{i,j,k}}{\partial x_r^d}=(-1)^{1-i}\cdot(x_g^d)^j(1-x_g^d)^{1-j}\cdot(x_b^d)^k(1-x_b^d)^{1-k},\\
    \frac{\partial V_{i,j,k}}{\partial x_g^d}=(x_r^d)^i(1-x_r^d)^{1-i}\cdot(-1)^{1-j}\cdot(x_b^d)^k(1-x_b^d)^{1-k},\\
    \frac{\partial V_{i,j,k}}{\partial x_b^d}=(x_r^d)^i(1-x_r^d)^{1-i}\cdot(x_g^d)^j(1-x_g^d)^{1-j}\cdot(-1)^{1-k},
\end{split}
\end{equation}
and,
\begin{equation}
\label{eq:backward-3}
\begin{split}
    \frac{\partial x_r^d}{\partial x_r^0}=-\frac{1-x_r^d}{x_r^1-x_r^0},\ \frac{\partial x_r^d}{\partial x_r^1}=-\frac{x_r^d}{x_r^1-x_r^0},\\
    \frac{\partial x_g^d}{\partial x_g^0}=-\frac{1-x_g^d}{x_g^1-x_g^0},\ \frac{\partial x_g^d}{\partial x_g^1}=-\frac{x_g^d}{x_g^1-x_g^0},\\
    \frac{\partial x_b^d}{\partial x_b^0}=-\frac{1-x_b^d}{x_b^1-x_b^0},\ \frac{\partial x_b^d}{\partial x_b^1}=-\frac{x_b^d}{x_b^1-x_b^0}.
\end{split}
\end{equation}
Note that since $x_c^0,x_c^1$ are elements in $\hat{P}$, \Cref{eq:backward-1,eq:backward-2,eq:backward-3} indeed define the partial derivatives with respect to $\hat{P}$, allowing the loss gradients to flow back to the sampling coordinates, and therefore back to the sampling intervals and any preceding neural network modules.
The AiLUT-Transform works on each pixel independently, so it is highly parallelizable, especially on GPUs.

\section{Experimental Details}
\label{sec:appl-impl}

Both datasets used in our work, i.e., the MIT-Adobe FiveK \cite{CVPR11FiveK} and PPR10K \cite{ICCV21PPR10K}, are publicly released, free of charge for research use, and contain no personally identifiable information. We conducted our experiments based on PyTorch 1.8.1 \cite{PyTorch} and the MMEditing toolbox (v0.11.0, under Apache 2.0 license) \footnote{\url{https://github.com/open-mmlab/mmediting}}. The proposed AiLUT-Transform is complied as a PyTorch CUDA extension using CUDA 10.2. In the following section, we provide more experimental details about these two datasets.

\subsection{FiveK}

\paragraph{Datasets Preprocessing}

Experiments on the FiveK dataset are conducted on two different resolutions (480p and 4K) and two different applications (retouching and tone mapping). For retouching and tone mapping on 480p, we directly download the dataset released by \cite{TPAMI203DLUT} \footnote{\url{https://github.com/HuiZeng/Image-Adaptive-3DLUT}} for model training and testing. To obtain the original 4K images, we follow \cite{TPAMI203DLUT} to download the original dataset \footnote{\url{https://data.csail.mit.edu/graphics/fivek}} and use the Adobe Lightroom software to pre-process the RAW images. We use the same training/testing splits as \cite{TPAMI203DLUT} to divide the dataset into 4500 training pairs and 500 testing pairs.
For the sake of research reproducibility, we provide the split files and the detailed instruction of data pre-processing at \url{https://github.com/ImCharlesY/AdaInt}.

We train all our models on the 480p resolution. During training, input images (8-bit sRGB for retouching and 16-bit CIE XYZ for tone mapping) are normalized to $[0,1]$ for unified processing and are randomly augmented to mitigate overfitting and facilitate performance. The augmentations include random ratio cropping, random horizontal flipping, and random color jittering. The trained models can be directly applied to the original 4K resolution without performance drop (see \iftoggle{selfappendix}{Table 2 in the paper}{\Cref{tab:sota-fivek-srgb}}). This protocol significantly speeds up the training stage and also demonstrates the flexibility and scalability of our method.

\begin{table}[t]
    \centering
    \begin{tabular}{ccc}
    \toprule[1pt]
    Id & Layer           & Output Feature Shape \\
    \midrule[1pt]
    0  & Bilinear Resize    & $3\times 256\times 256$ \\
    1  & Conv3x3, LeakyReLU & $16\times 128\times 128$ \\
    2  & InstanceNorm       & $16\times 128\times 128$ \\
    3  & Conv3x3, LeakyReLU & $32\times 64\times 64$ \\
    4  & InstanceNorm       & $32\times 64\times 64$ \\
    5  & Conv3x3, LeakyReLU & $64\times 32\times 32$ \\
    6  & InstanceNorm       & $64\times 32\times 32$ \\
    7  & Conv3x3, LeakyReLU & $128\times 16\times 16$ \\
    8  & InstanceNorm       & $128\times 16\times 16$ \\
    9  & Conv3x3, LeakyReLU & $128\times 8\times 8$ \\
    10 & Dropout (0.5)      & $128\times 8\times 8$ \\
    11 & AveragePooling     & $128\times 2\times 2$ \\
    12 & Reshape            & $512$ \\
    \bottomrule[1pt]
    \end{tabular}
    \caption{Network architecture of the backbone module (the mapping $f$ in the paper) on the FiveK \cite{CVPR11FiveK} dataset.}
    \label{tab:arch-backbone}
    \vspace{-0.4cm}
\end{table}

\paragraph{Model Architectures}
\label{sec:suppl-arch}

The model architectures employed on the FiveK experiments are listed in \Cref{tab:arch-backbone,tab:arch-lut-generator,tab:arch-adaint}. For the backbone network (mapping $f$), we directly adopt the 5-layer \footnote{The term "5-layer" corresponds to 5 convolutional layers.} backbone in \cite{TPAMI203DLUT}. The mapping $h$ uses a cascade of two fully-connected (FC) layers, which reformulates the implementation in \cite{TPAMI203DLUT}, \textit{i.e.}, the first FC layer (mapping $h_0$) is responsible for predicting $M$ input-dependent fusing weights, whereas the second FC layer (mapping $h_1$) encodes the parameters of $M$ basis 3D LUT. For the instantiation of mapping $g$, a single FC layer is adopted. We follow \cite{TPAMI203DLUT} to initialize the parameters of $f$ and $h$. As for the proposed AdaInt module (mapping $g$), we initialize its weights and bias to all 0s and 1s, which makes the predicted sampling intervals start from a uniform state, hence stabilizing the training of AdaInt.

\vspace{-0.2cm}

\paragraph{Training Details}
We train our models for 400 epochs with a fixed learning rate of $1\times 10^{-4}$ using the standard Adam optimizer \cite{Adam}. The mini-batch size is set to 1. To further stabilize the AdaInt learning and facilitate the learning of more output color values in the 3D lattice, we decay the learning rate of $g$ by a factor of 0.1 and freeze its parameters in the first 5 training epochs.

\begin{table}[t]
    \centering
    \begin{tabular}{ccc}
    \toprule[1pt]
    Id & Layer & Output Feature Shape \\
    \midrule[1pt]
    0 & FC & $M$ \\
    1 & FC & $3N_s^3$ \\
    2 & Reshape & $3\times N_s\times N_s\times N_s$ \\
    \bottomrule[1pt]
    \end{tabular}
    \caption{Network architecture of the mapping $h$ in the paper. "FC" denotes the fully-connected layer.}
    \label{tab:arch-lut-generator}
    \vspace{-0.2cm}
\end{table}

\begin{table}[t]
    \centering
    \begin{tabular}{ccc}
    \toprule[1pt]
    Id & Layer & Output Feature Shape \\
    \midrule[1pt]
    0 & FC & $3(N_s-1)$ \\
    1 & Reshape & $3\times (N_s-1)$ \\
    2 & Softmax & $3\times (N_s-1)$ \\
    3 & Cumsum & $3\times (N_s-1)$ \\
    4 & ZeroPad & $3\times N_s$ \\
    \bottomrule[1pt]
    \end{tabular}
    \caption{Network architecture of the mapping $g$ in the paper. "FC" denotes the fully-connected layer, and "Cumsum" is the operation to conduct cumulative summation.}
    \label{tab:arch-adaint}
    \vspace{-0.4cm}
\end{table}

\subsection{PPR10K}

\paragraph{Datasets Preprocessing}

Experiments on the PPR10K dataset are conducted on the 360p resolution for the photo retouching application. The official training/testing ($8,875:2,286$) splits are adopted. During training, images augmented by the dataset creator are used as our inputs. Please refer to \cite{ICCV21PPR10K} for more details. Besides the pre-augmentations, we further apply some commonly-used data augmentation methods such as random ratio cropping and random horizontal flipping. Since the dataset is relatively larger-scale than the FiveK dataset, images are also resized into a unified $448\times 448$ resolution to enable mini-batch processing for speeding up the training stage.

\vspace{-0.2cm}

\paragraph{Model Architectures}

For experiments on the PPR10K dataset, the mapping $g$ and $h$ share identical architectures (\Cref{tab:arch-adaint,tab:arch-lut-generator}) with those on the FiveK dataset. For the backbone network (the mapping $f$), we follow \cite{ICCV21PPR10K} to adopt the ResNet-18 \cite{ResNet} (with a preceding bilinear layer resizing images into $224\times 224$ resolution) for a fair comparison. The initializations of $g$ and $h$ are the same as those in the FiveK experiments, whereas the ResNet-18 backbone is initialized using ImageNet \cite{ImageNet} pretrained weights.

\vspace{-0.2cm}

\paragraph{Training Details}
We train our models for 200 epochs with a mini-batch size of 16 to speed up the training stage on such a large-scale dataset. For other settings, they are kept consistent with those in the FiveK experiments.

\begin{figure*}
    \centering
    \includegraphics[width=\linewidth]{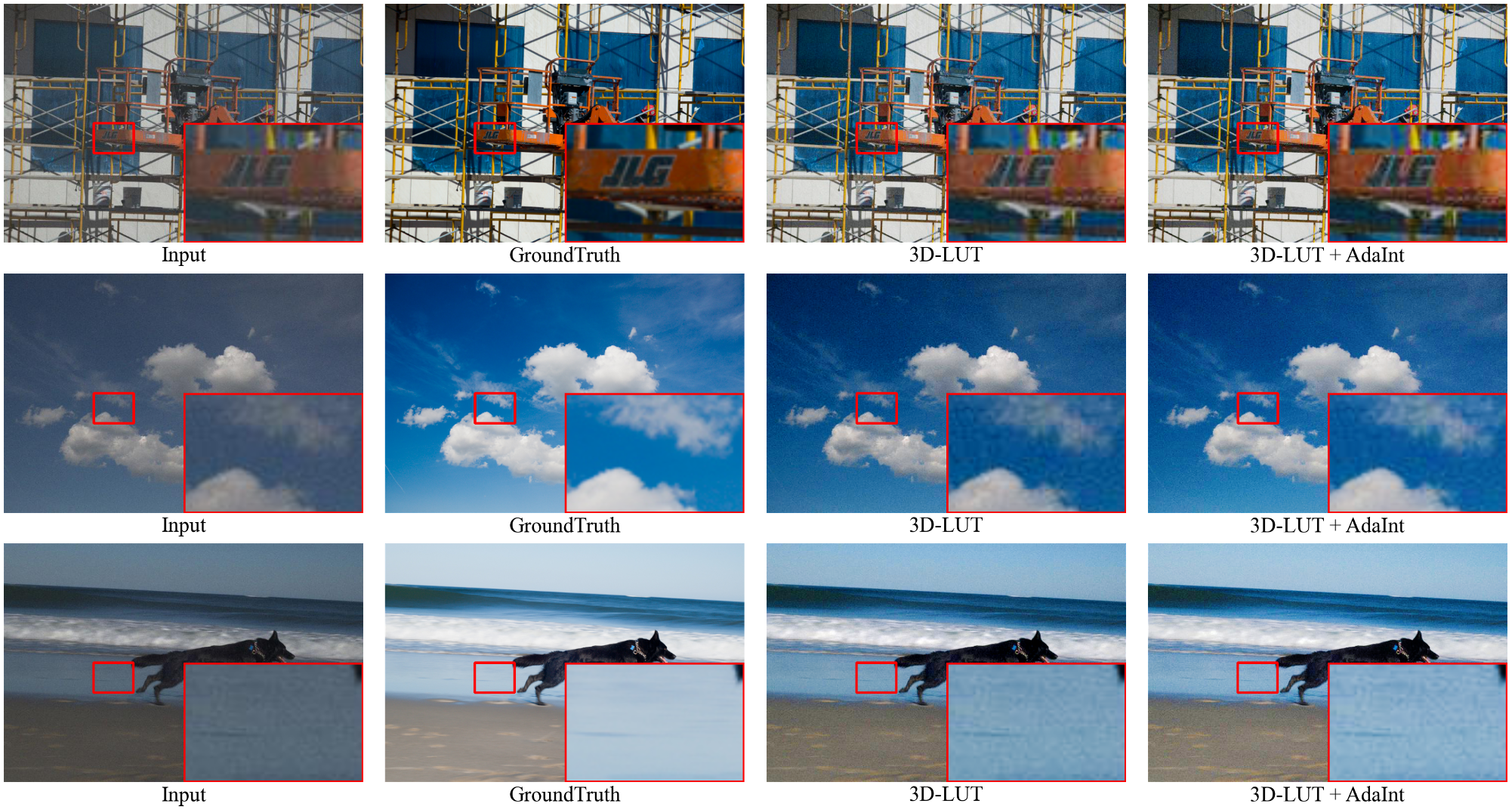}
    \caption{Qualitative results on noisy images for \textbf{photo retouching} on the \textbf{FiveK} dataset (480p)~\cite{TPAMI203DLUT}. Best viewed on screen.}
    \label{fig:noise}
    \vspace{-0.4cm}
\end{figure*}

\section{More Visualization of AdaInt}
\label{sec:appl-adaint}

To better investigate the behavior of AdaInt, inspired by \cite{ECCV12Nonuniform}, we introduce the per-color-channel \textit{Accumulative Error Histogram (AEH)}, which, to some extent, can indicate locations in the color ranges where more dense sampling points are required.

\begin{figure*}
    \centering
    \includegraphics[width=\linewidth]{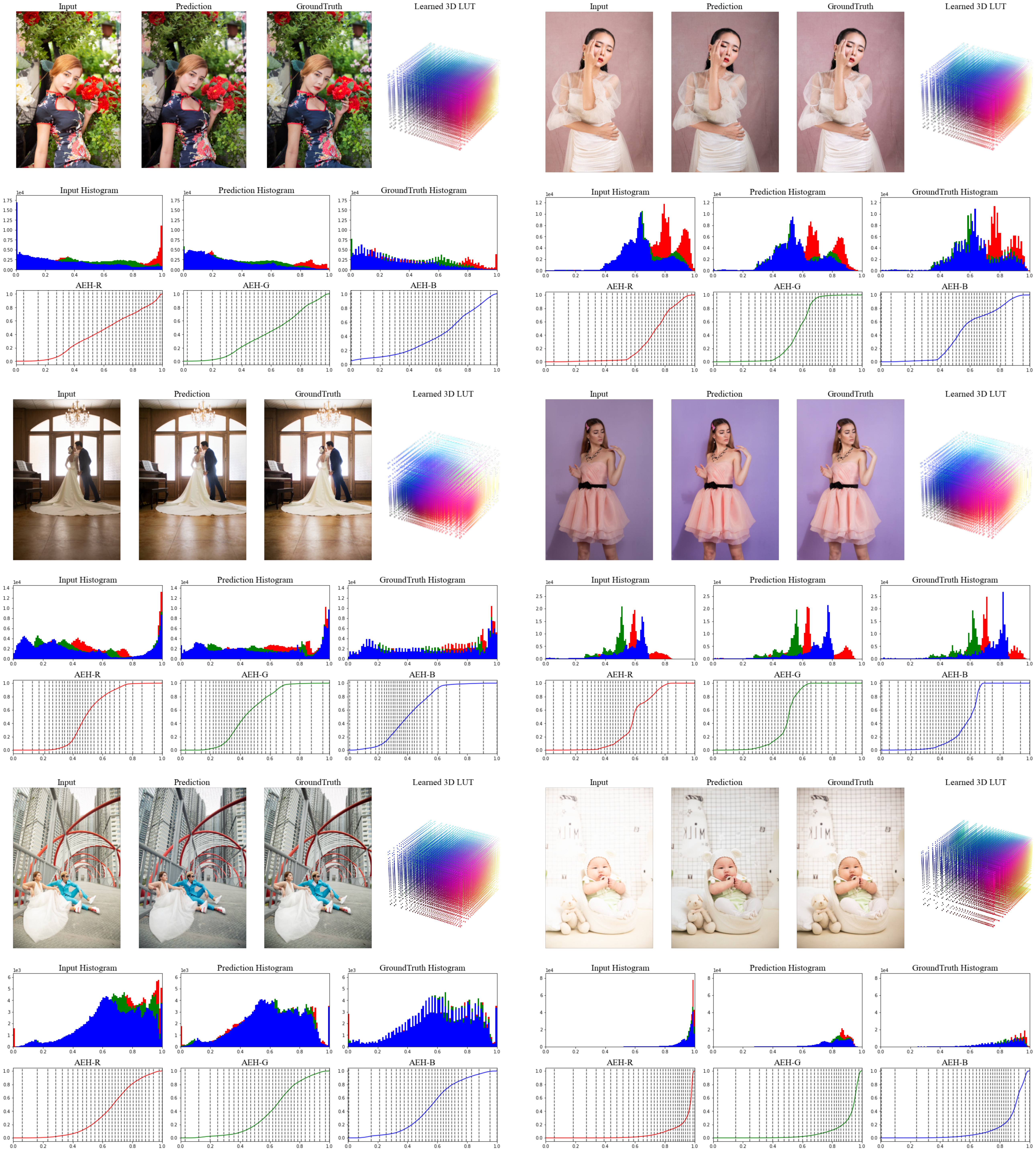}
    \caption{Illustration of the learned sampling coordinates and the corresponding 3D LUTs for \textbf{photo retouching} on the \textbf{PPR10K} dataset (360p)~\cite{ICCV21PPR10K}. The bottom row visualizes the learned sampling coordinates on the so-called per-color-channel \textit{Accumulative Error Histogram (AEH)}~\cite{ECCV12Nonuniform}. The regions in the AEH exhibiting high curvature indicate wherein more sampling points are needed. Best viewed on screen.}
    \label{fig:adaint-1}
\end{figure*}

The AEH bins the intensity changes between input and target images according to the values of the input pixels. Therefore, it indicates the intensity change required for each input pixel value to transform the image. Given an input image $X\in[0,1]^{3\times H\times W}$ and the corresponding groundtruth image $Y\in[0,1]^{3\times H\times W}$, the AEH is constructed by first calculating the error map $D$ between $X$ and $Y$:
\begin{equation}
    D_{c,h,w} = (X_{c,h,w}-Y_{c,h,w})^2,
\end{equation}
where $c=\{r,g,b\}$, $h\in\mathbb{I}_0^{H-1}$, and $w\in\mathbb{I}_0^{W-1}$. Before computing the histogram, we divide the input color range $[0,1]$ for a single channel uniformly into $N_\text{bin}$ bins and assign each input pixel into one of them according to its value at the corresponding color channel:
\begin{equation}
    \Omega_c^k=\{X_{c,h,w}| k\Delta\leq X_{c,h,w}<(k+1)\Delta\},
\end{equation}
where $\Delta=1/N_\text{bin}$ and $\Omega_c^k$ is the $k$-th bin ($k\in\mathbb{I}_0^{N_\text{bin}-1}$) for channel $c$. The normalized error histogram $H_c$ for channel $c$ is then computed as:
\begin{equation}
    H_c[k] = \frac{1}{\Theta_c}\sum_{X_{c,h,w}\in\Omega_c^k}D_{c,h,w},
\end{equation}
where $\Theta_c$ is a normalization term such that $\sum_k H_c[k] = 1$. The final per-color-channel AEH can be easily derived by applying accumulative summation on $H_c$. In this work, we set $N_\text{bin}=1000$. The AEH can be visualized as a 1D curve as shown in \Cref{fig:adaint-1}. The partial curve exhibiting higher curvature shows the color range requires a more substantial intensity change. Though not precise, the AEH indicates regions in the color range where the non-linearity is most prominent, and hence more sampling points are required.

We provide more visualization in \Cref{fig:adaint-1} of the learned sampling coordinates and the corresponding 3D LUTs for different images from the PPR10K dataset. It is clearly shown that the learned sampling coordinates are inclined to locate densely in the color range where the AEH exhibits high curvature, which suggests that our AdaInt is able to capture the complexity of the underlying optimal color transform with the adaption to the image content and achieve a more optimal sampling point allocation.

\section{Additional Qualitative Comparisons}
\label{sec:appl-quality}

\begin{figure*}[ht]
    \centering
    \includegraphics[width=0.8\linewidth]{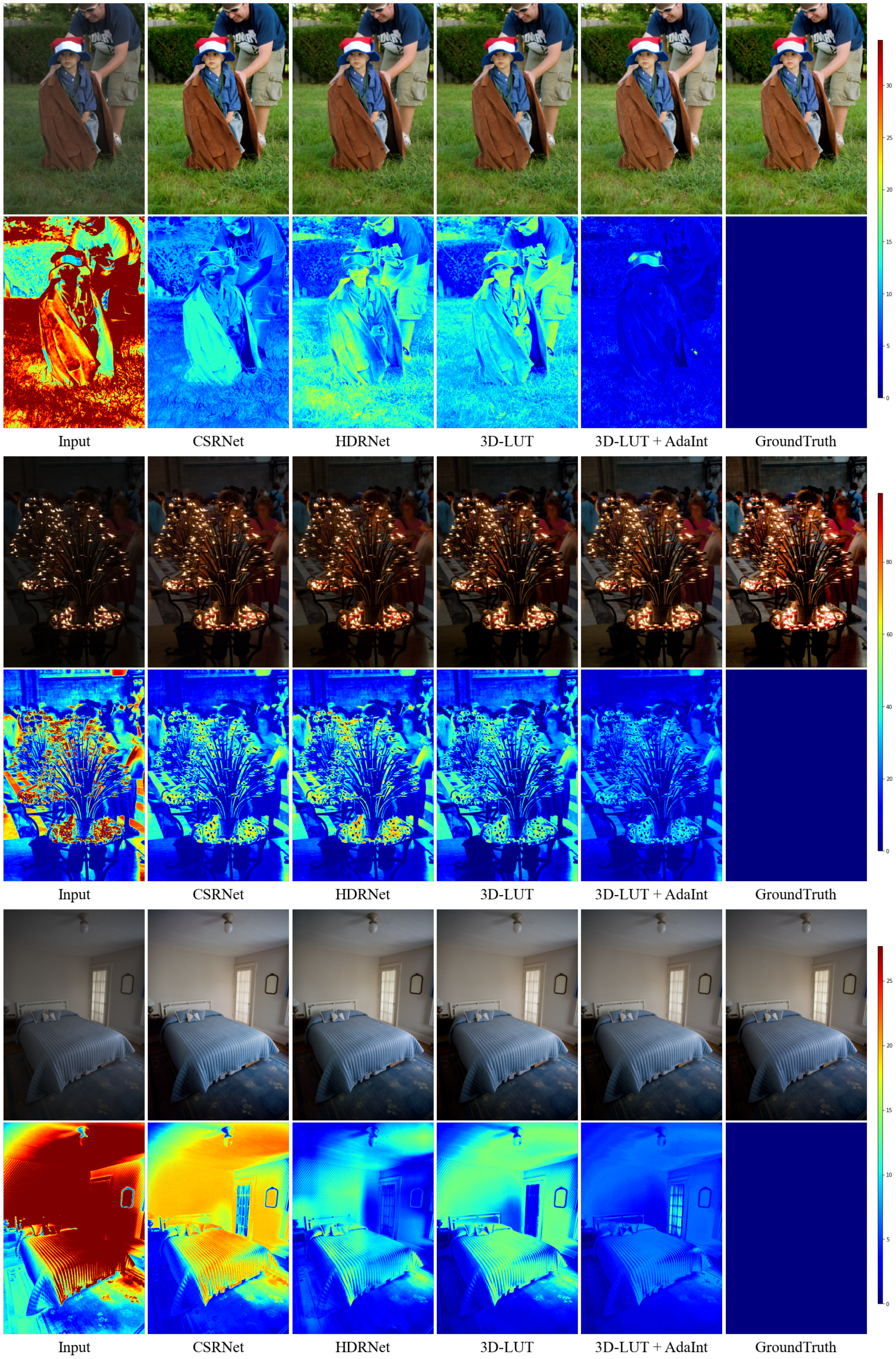}
    \caption{Additional qualitative comparisons of different methods for \textbf{photo retouching} on the \textbf{FiveK} dataset (4K)~\cite{CVPR11FiveK}, and the corresponding error maps. Best viewed in zoom in.}
    \label{fig:sota-fivek-1}
\end{figure*}

\begin{sidewaysfigure*}[ht]
    \centering
    \includegraphics[width=\linewidth]{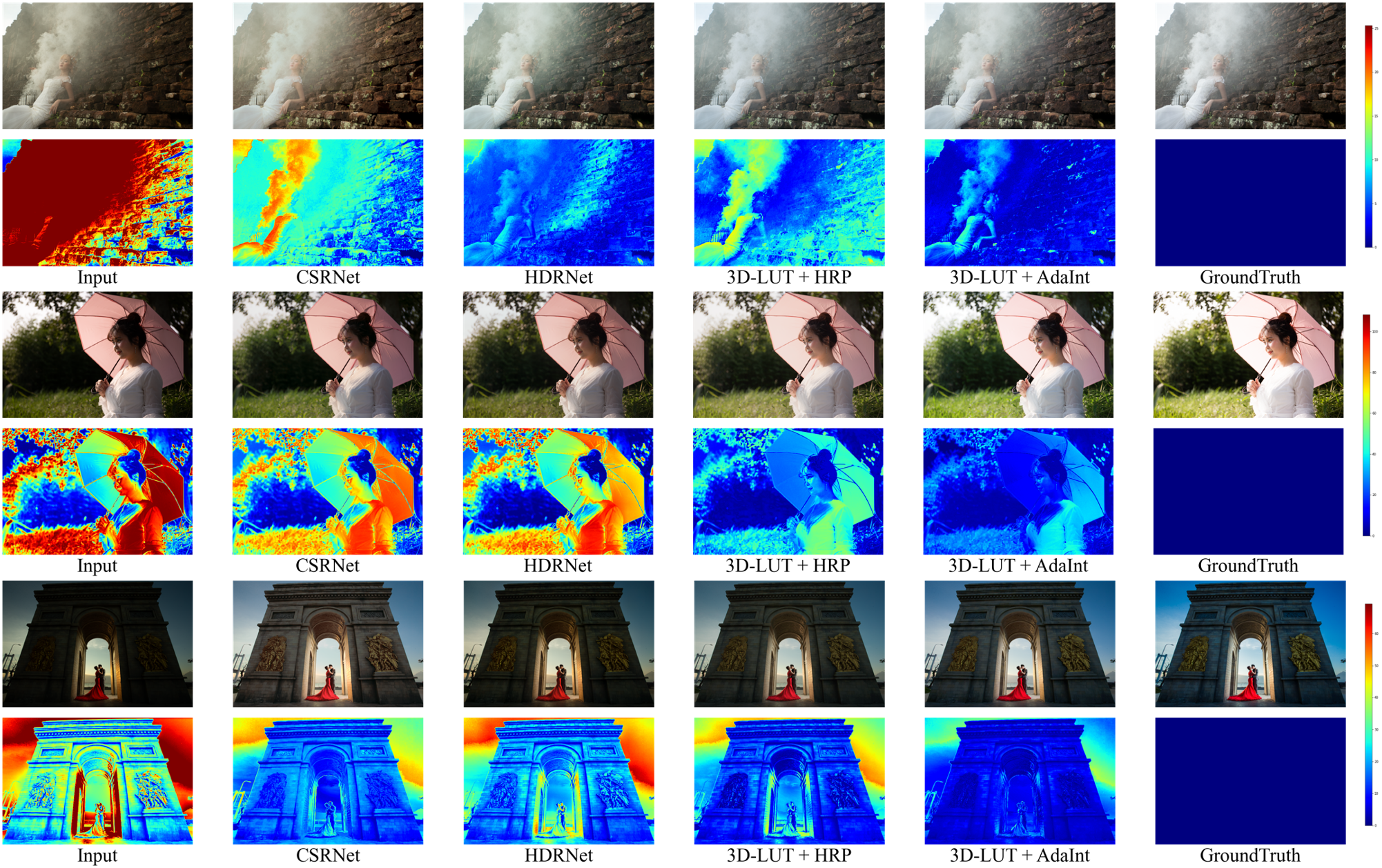}
    \caption{Additional qualitative comparisons of different methods for \textbf{photo retouching} on the \textbf{PPR10K} dataset (360p)~\cite{CVPR11FiveK}, and the corresponding error maps. Best viewed in zoom in.}
    \label{fig:sota-ppr10k-1}
\end{sidewaysfigure*}

In this section, we provide additional visual comparisons on the FiveK (4K) dataset in \Cref{fig:sota-fivek-1} and on the PPR10K (360p) dataset in \Cref{fig:sota-ppr10k-1}.

\section{Real-time Performance on 8K Images}

We also follow the protocol in \cite{TPAMI203DLUT} to resize the images into 8K (7680$\times$4320) resolution and measure the GPU inference time of both the baseline \cite{TPAMI203DLUT} and our method on a V100 GPU. The baseline can execute at a speed of 2.36 ms per 8K image, while AdaInt only increases the runtime to 2.54 ms, which still exceeds the requirement of real-time processing by a large margin. Note that the high efficiency of our method mainly owns to the characteristic that the computational cost of the CNN network is fixed as it operates on a downsampled, fixed-resolution version of the input, and the proposed AiLUT-Transform can be highly parallelized via customized CUDA code.

\section{Limitation}

As discussed in \iftoggle{selfappendix}{Section 5 in the paper}{\Cref{sec:conclusion}}, the proposed method is based on pixel-wise mapping and thus may fail to tackle heavy noise existing in the input. For verification, we randomly select three images from the FiveK dataset \cite{CVPR11FiveK} and add Gaussian noise (with a standard deviation of $\sigma=0.02$) to them. As shown in \Cref{fig:noise}, both the baseline \cite{TPAMI203DLUT} and our method could not eliminate the added noise due to the lack of filtering capability. Constructing pixel-wise LUTs \cite{ICCV21SALUT} might be a possible solution but still requires careful investigation to prevent substantial increases in memory and computational costs. 

\section{Broader Impacts}
\label{sec:appl-limit}

The proposed method mimics the retouching style of human annotators based on learned statistics of the training dataset and as such will reflect biases in those data, including ones with negative societal impacts. Besides, there is no guarantee that the transformed colors by the method will please everybody due to the subjectiveness of the image/photo enhancement task. A possible mitigation strategy is to concern the preference of specific downstream users and retrains the method accordingly.

\end{document}